\documentclass[smallextended]{svjour3}       
\smartqed 
\usepackage{graphicx}
\usepackage{enumitem}
\usepackage{lscape}
\usepackage{tikz}
\usepackage{tikz-qtree}
\usetikzlibrary{calc,trees,positioning,arrows,chains}
\usepackage{epstopdf}
\usepackage{algorithm,algorithmic}
\usepackage{pgfplots}
\usepackage{subcaption}
\usepackage{mathtools}
\usepackage{hyperref} % [hidelinks]
\usepackage{multirow,multicol}
\usepackage{amssymb}
\usepackage{arydshln} 
\newsavebox{\measurebox}
\begin{document}

\title{Improving Predictions of Tail-end Labels using Concatenated BioMed-Transformers for Long Medical Documents}

\titlerunning{{Concatenated BioMed-Transformers for Tail-end Labels}}   

\author{Vithya~Yogarajan \and Bernhard~Pfahringer \and Tony~Smith \and Jacob~Montiel  }

\authorrunning{Yogarajan et al.} % if too long for running head

\institute{Vithya Yogarajan \at
              The University of Auckland.
              \email{vithya.yogarajan@auckland.ac.nz}           %  \\
           \and
           Jacob Montiel, Tony Smith, Bernhard Pfahringer  \at
          The University of Waikato.  \email{bernhard.pfahringer@waikato.ac.nz}  \\
}
 %         Vithya Yogarajan {0000-0002-6054-9543}, Jacob Montiel {0000-0003-2245-0718}, \\ Tony Smith {0000-0003-0403-7073}, 
%          Bernhard Pfahringer {0000-0002-3732-5787}

\date{Received: date / Accepted: date}

\maketitle

\begin{abstract}

Multi-label learning predicts a subset of labels from a given label set for an unseen instance while considering label correlations.  A known challenge with multi-label classification is the long-tailed distribution of labels. Many studies focus on improving the overall predictions of the model and thus do not prioritise tail-end labels. Improving the tail-end label predictions in multi-label classifications of medical text enables the potential to understand patients better and improve care. The knowledge gained by one or more infrequent labels can impact the cause of medical decisions and treatment plans.  This research presents variations of concatenated domain-specific language models, including multi-BioMed-Transformers, to achieve two primary goals. First, to improve F1 scores of infrequent labels across multi-label problems, especially with long-tail labels; second, to handle long medical text and multi-sourced electronic health records (EHRs), a challenging task for standard transformers designed to work on short input sequences. A vital contribution of this research is new state-of-the-art (SOTA) results obtained using TransformerXL for predicting medical codes. A variety of experiments are performed on the Medical Information Mart for Intensive Care (MIMIC-III) database. Results show that concatenated BioMed-Transformers outperform standard transformers in terms of overall micro and macro F1 scores and individual F1 scores of tail-end labels, while incurring lower training times than existing transformer-based solutions for long input sequences.  

\end{abstract}
\section{Introduction}

\begin{figure}[t]
    \centering
    \includegraphics[width=0.46\textwidth,height=2.2cm]{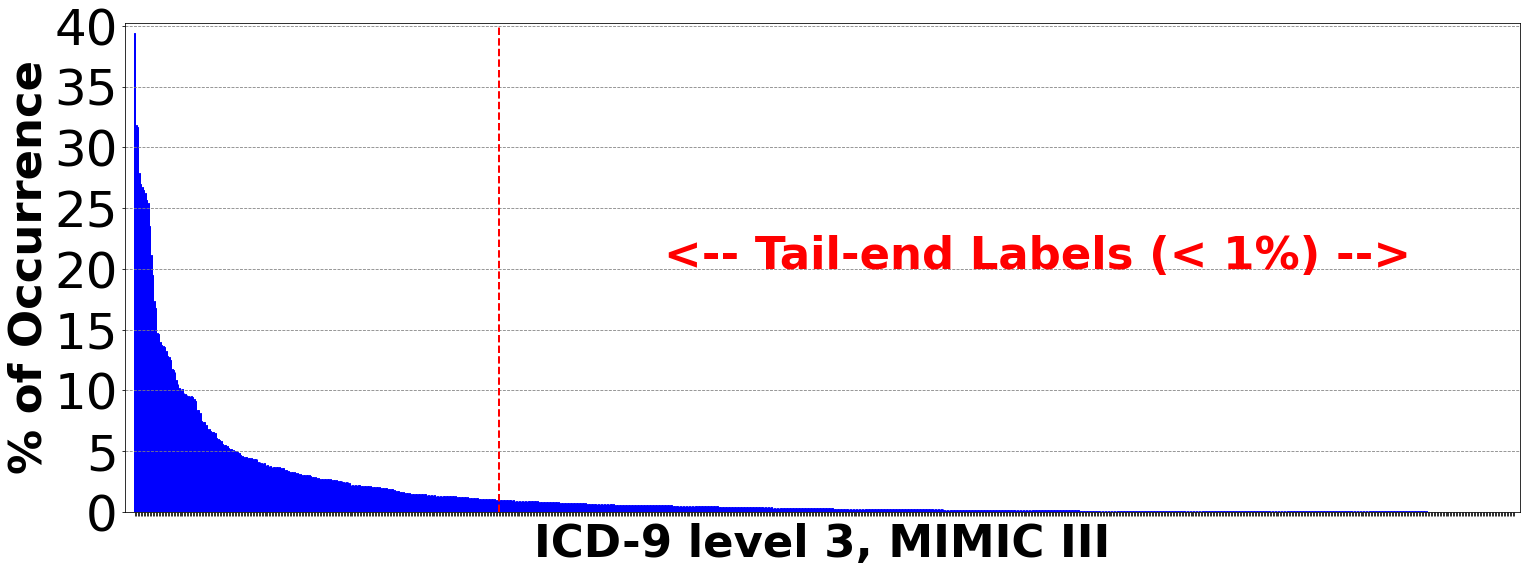}
    \hfill
    \includegraphics[width=0.46\textwidth,height=2.18cm]{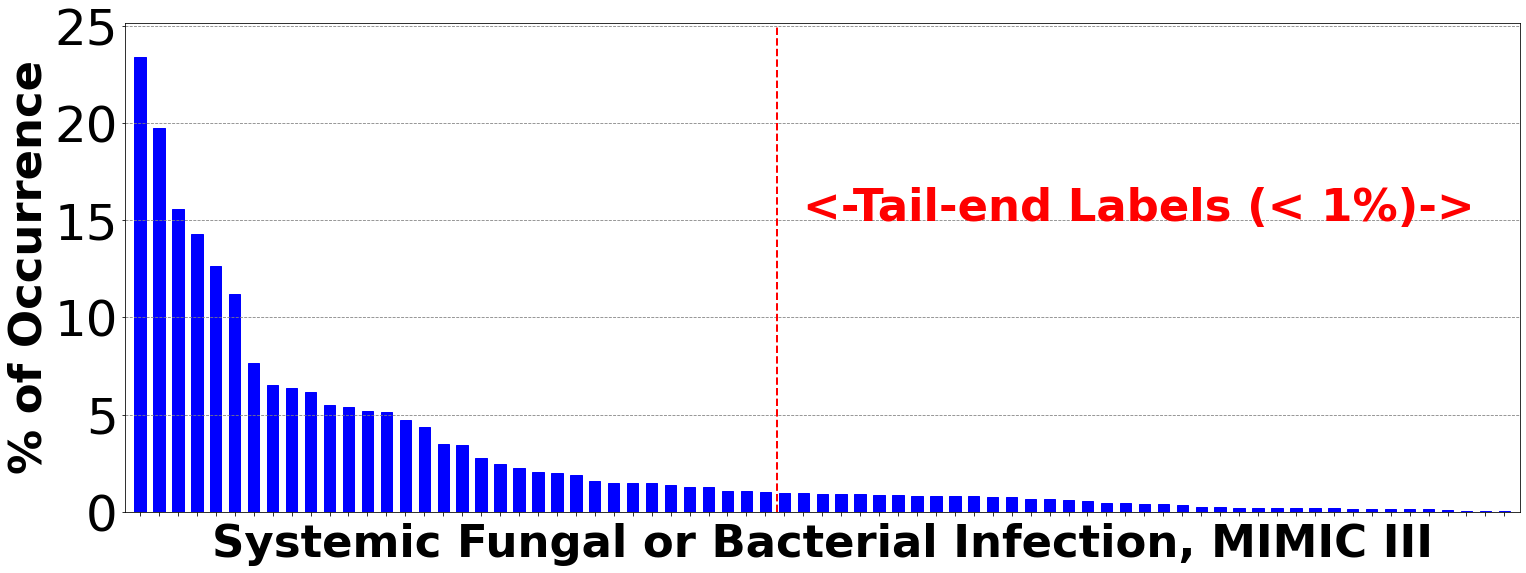}
    \caption{Percentage frequency of labels for ICD-9 level-3 codes with 923 labels (left) and systemic fungal or bacterial infection with 73 labels (right) for MIMIC-III data. The labels are ordered from most frequent (left) to least frequent (right) for each plot. The threshold for tail-end labels with \% Freq of occurrences $< 1\%$ is indicated for reference. }
    \label{fig:freq}
    \vspace{-1.5em}
\end{figure}

Multi-label text classification techniques enable predictions of treatable risk factors in patients, aiding in better life expectancy and quality of life~\cite{aubert2019patterns}. The goal of multi-label learning is to predict a subset of labels for an unseen instance from a given label set while considering label correlations~\cite{zhang2013review}. One of the known challenges with multi-label classification is the long-tailed distribution of labels. In general, with multi-label problems, a small subset of the labels are associated with a large number of instances, and a significant fraction of the labels are associated with a small number of instances (as shown in Figure~\ref{fig:freq}). 

There are some examples of studies that focus on exploiting label structure~\cite{zhang2018deep} and label co-occurrence patterns~\cite{kurata2016improved}. However, in studies especially relating to medical text, the focus is on improving the overall performance of the model instead of individual tail-end labels~\cite{moons2020comparison,amin2019mlt}. There are also examples of studies, such as Wei and Li (2019)~\cite{wei2019does}, which demonstrate that tail-end labels have minimal impact on the overall performance. However, prediction of infrequent labels in order to understand all aspects of a patient's prognosis is as crucial as predicting frequent labels~\cite{flegel2018we}. The knowledge gained by one or more infrequent labels can impact the cause of medical decisions, treatment plans and patient care.   

This research explores the opportunity to improve predictions of tail-end labels using transformers for medical-domain specific tasks by exploiting models pre-trained on health data. We consider the option of using three variations of concatenated language models: multi-CNNText, multi-BioMed-Transformers and CNNText with Transformers. We show concatenated BioMed-Transformers improve tail-end predictions compared to other neural networks and single transformers.    

In addition to improving the tail-end performance, we demonstrate concatenated domain-specific transformer models are a solution for handling text data with extended text and multi-sources of texts. For short or truncated electronic health records (EHRs), medical domain-specific transformer models outperform state-of-the-art (SOTA) methods for many classification tasks, including predicting medical codes and name entity recognition~\cite{yogarajan2021trans,domains,gu2020domain}. However, given that most transformer models are limited to a maximum sequence length of 512 tokens, with some exceptions, there is still a gap in alternative solutions for long documents. Transformer models such as Longformer~\cite{beltagy1904longformer} and TransformerXL~\cite{dai2019transformer} can handle longer sequences and perform better than other language models for long documents. Unfortunately, these models require considerable amounts of memory and processing time. In contrast, concatenated domain-specific transformers require fewer resources. 

We  also present new SOTA results using TransformerXL for predicting medical codes. We compare these results directly with the most recent (Nov, 2021) published SOTA~\cite{liueffective} for the exact same multi-label text classification problem. 

We compare concatenated domain-specific transformer models with standard language models for increasingly larger multi-label problems with 30, 42, 50, 73, 158 and 923 labels. The multi-label problems considered in this paper are: predicting ICD-9 codes for ICD-9 hierarchy levels, most frequent 50 ICD-9 codes, cardiovascular disease, COVID-19 patient shielding (introduced in Yogarajan et al (2021)~\cite{yogarajan2021predicting}) and systemic fungal or bacterial disease. 
The contributions of this work are:
\begin{enumerate}
    \item analyse the effectiveness of using concatenated domain-specific language models, multi-CNNText, multi-BioMed-Transformers and CNNText with Transformers, for predicting medical codes from EHRs for multiple document lengths, multi-sources of texts and number of labels;
    \item show that concatenated domain-specific transformers improve F1 scores of infrequent labels; 
     \item show improvements in overall micro and macro F1 scores and achieve such improvements with fewer resources;
     \item present new SOTA results for predicting medical codes from EHRs. 
\end{enumerate}

\section{Related Work}

In the last two to three years, there have been considerable advancements in transformer models, which have shown substantial improvements in many NLP tasks, including BioNLP tasks~\cite{gu2020domain,yang2020clinical}. With minimum effort, transfer learning of pre-trained models by fine-tuning on downstream supervised tasks achieves very good results~\cite{amin2020exploring,amin2019mlt}. Examples of BioNLP tasks where transformers have shown performance improvements include named entity recognition, question answering, relation extraction, and clinical concept extraction tasks~\cite{gu2020domain,yang2020clinical,domains}. 

A significant obstacle for transformers is the 512 token size limit they impose on input sequences~\cite{9364676}. Gao et al. (2021)~\cite{9364676} presents evidence showing BERT-based models under-perform in clinical text classification tasks with long input data, such as MIMIC-III~\cite{johnson2016mimic}, when compared to a CNN trained on word embeddings that can process the complete input sequences. Si and Roberts (2021)~\cite{si2021hierarchical} presents an alternative system to overcome the issue of long documents, where transformer-based encoders are used to learn from words to sentences, sentences to notes and notes to patients progressively. This transformer-based hierarchical attention networks system presents SOTA methods for in-hospital mortality prediction and phenotype predictions using MIMIC-III. However, it requires considerable computational resources~\cite{si2021hierarchical}. Chalkidis et al. (2020)~\cite{chalkidis2020empirical} proposes a similar hierarchical version using SCI-BERT to deal with long documents for predicting medical codes from MIMIC III. Here SCI-BERT reads words of each sentence, resulting in sentence embeddings. This is followed by a self-attention mechanism that reads the sentence embeddings to produce single document embeddings fed through an output layer. Unfortunately, HIER-SCI-BERT performed poorly compared to other neural networks~\cite{chalkidis2020empirical}. One possible reason for poor results is the use of a continuously pre-trained BERT model ~\cite{chalkidis2020empirical}. The continuous training approach would initialise with the standard BERT model, pre-trained using Wikipedia and BookCorpus. It then continues the pre-training process with a masked language model and next-sentence prediction using domain-specific data. In this case, the vocabulary is the same as the original BERT model, which is considered a disadvantage for domain-specific tasks~\cite{gu2020domain}. For our research, PubMedBERT~\cite{gu2020domain}, a domain-specific BERT based model trained solely on biomedical text, is used. 

Our research focuses on automatically predicting medical codes from medical text as the multi-label classification task. Examples of predicting medical codes using transformers include ICD-10 predictions from German documents~\cite{amin2019mlt,sanger2019classifying}, and predicting frequent medical codes from MIMIC-III~\cite{biswas2021transicd,yogarajan2021trans}. These examples restrict themselves to (1) truncated text sequences of $<512$ tokens and (2) predicting frequent labels~\cite{biswas2021transicd,amin2020exploring}. MIMIC-III consists of many infrequent labels, as shown in Figure~\ref{fig:freq}, where most codes only occur in a small number of clinical documents. This research focuses on improving the predictive accuracy for infrequent labels and using long medical texts. Moons et al. (2020)~\cite{moons2020comparison} presents a survey of deep learning methods for ICD coding of medical documents and indicates Convolutional Attention for Multi-Label classification (CAML)~\cite{mullenbach2018explainable} as the SOTA method for automatically predicting medical codes from EHRs.   Yogarajan et al. (2021)~\cite{yogarajan2021trans} presents evidence to show that domain-specific transformers outperform CAML for truncated sequences. Liu et al (2021)~\cite{liueffective} presents the most recent evidence where EffectiveCAN --an effective convolution attention network-- outperforms SOTA for predicting medical codes. We extend the findings in Yogarajan et al. (2021)~\cite{yogarajan2021trans} by providing evidence to show TransformerXL outperforms CAML and sets new SOTA results for predicting medical codes. We also present a direct comparison with EffectiveCAN for the same multi-label problem with the same labels and data to show transformers such as TransformerXL outperform SOTA.   

 \vspace{-1.5em}
\section{Data}\label{sec:data}

Medical Information Mart for Intensive Care (MIMIC-III) is one of the most extensive publicly available medical databases \cite{johnson2016mimic,goldberger2000physiobank} with more than 50,000 patient EHRs. It contains data including billing, laboratory, medications, notes, physiological information, and reports. Among the available free-form medical text, more than 90\% of the unique hospital admissions contain at least one discharge summary (\textsf{dis}). In addition to the free-form medical text from \textsf{dis}, this research also makes use of text summary of categories ECG (\textsf{ecg}) and Radiology(\textsf{rad}). As with most free form EHRs, MIMIC-III text data includes acronyms, abbreviations, and spelling errors. For example (data as presented in MIMIC III with errors):
\begin{quote}
    \textit{82 yo M with h/o CHF, COPD on 5 L oxygen at baseline, tracheobronchomalacia s/p stent, presents with acute dyspnea over several days, and lethargy...}
\end{quote}

MIMIC-III data includes long documents, where \textsf{dis} ranges from 60 to 9,500 tokens with an average of 1,513 tokens and \textsf{rad} with an average of 2,500 tokens. The document lengths of \textsf{ecg} are short with an average of 84 tokens. In this research, MIMIC-III text is pre-processed by removing tokens that contain non-alphabetic characters, including all special characters and tokens that appear in less than three training documents. 

The discharge summary is split into equal segments for a given hospital admission, and each section is labelled text $1,...,4$. For example, for two splits, if a given discharge summary is $700$ tokens long, text 1 is the first 350 tokens, and text 2 is the last 350 tokens. In the case of a lengthy document, if the discharge summary is $2500$ tokens long, text 1 is the first $1,250$ tokens, and text 2 is the last $1,250$ tokens. For multi-BioMed-Transformers where the maximum sequence length is $512$, each of text $1,...,4$ is truncated to $512$ tokens. There are many other ways to split the text, including sequential splits. For instance, with the first example above, text 1 being the first $512$ tokens, and text 2 being the remainder $238$ tokens. Each of these decisions has some advantages and disadvantages. After preliminary experiments, the decision was made to split the discharge summary into equal sections. 
This research presents results for the following configurations:
\begin{enumerate}\addtocounter{enumi}{-1}
    \item \textsf{dis$_{1 \text{ of } 2}$} + \textsf{dis$_{2 \text{ of } 2}$}
     \item \textsf{dis$_{1 \text{ of } 3}$} + \textsf{dis$_{2 \text{ of } 3}$} + \textsf{dis$_{3 \text{ of } 3}$}.
   \item \textsf{dis$_{1 \text{ of } 2}$} + \textsf{dis$_{2 \text{ of } 2}$} + \textsf{ecg}. 
   \item \textsf{dis$_{1 \text{ of } 2}$} + \textsf{dis$_{2 \text{ of } 2}$}  + \textsf{rad}. 
    \item \textsf{dis$_{1 \text{ of } 2}$} + \textsf{dis$_{2 \text{ of } 2}$} + \textsf{ecg} + \textsf{rad}. 
   
    \item \textsf{dis$_{1 \text{ of } 4}$} + \textsf{dis$_{2 \text{ of } 4}$} + \textsf{dis$_{3 \text{ of } 4}$} + \textsf{dis$_{4 \text{ of } 4}$}.
     \item \textsf{dis}   + \textsf{ecg}. 
    \item \textsf{dis}   + \textsf{rad}. 
\end{enumerate}

\section{Multi-label Datasets and Labels}

We consider predicting ICD-9 codes (standards for international Statistical Classification of Diseases and Related Health Problems) from EHRs as flat multi-label problems. ICD codes are used to classify diseases, symptoms, signs, and causes of diseases. Almost all health conditions can be assigned a unique code. Manual assigning of medical codes requires expert knowledge and is very time-consuming. Thus, the ability to predict and automate medical coding is vital. ICD-9 codes are grouped in a hierarchical tree-like structure by the World Health Organisation. In this research, we focus on levels 2 and 3 for MIMIC-III data containing 158 labels at level 2 and 923 labels at level 3 with associated medical text for the patient. In addition, we consider case studies, cardiovascular disease, COVID-19 patient shielding, and systemic fungal or bacterial infections, where commonly used medical codes are used as labels. As mentioned earlier, for the purposes of direct comparison with the recently published SOTA, the most frequent 50 ICD-9 codes in MIMIC-III are also considered.

\begin{table}[t!]
    \centering
    \caption{Statistics of multi-label classification problems. Counts for frequent and infrequent, or tail-end labels, are also provided. * MIMIC III Top50 is the most frequent 50 labels, hence no tail labels, and is only used in this research for direct SOTA comparison.  }
    \label{tab:lab_den}
    \begin{tabular}{l@{\hspace{.2cm}}r@{\hspace{.2cm}}r@{\hspace{.2cm}}r@{\hspace{.2cm}}r@{\hspace{.2cm}}r@{\hspace{.2cm}}r}
        \hline
         \textbf{Multi-label Problems}  & \textbf{q}  & \textbf{\# Inst} & \textbf{LCard}  & \textbf{LDens} & {\textbf{LFreq}} $\geq 1\%$ & {\textbf{LFreq}} $< 1\%$ \\ 
        \hline
        MIMIC-III Level 3  & 923 & 52,722 & 14.43 &0.02 &244 & 679 \\ 
        MIMIC-III Level 2  & 158 & 52,722 & 11.61 &0.07 & 100 & 58  \\ 
        MIMIC-III Top50* & 50 & 50,957 & 5.60 & 0.11 & 50 & 0 \\
        Fungal or bacterial  & 73 & 30,814 &2.06 & 0.03  &34 & 39 \\ 
        COVID-19~\cite{yogarajan2021predicting}  & 42&35,458  &1.84 & 0.04 & 27 & 15 \\
        Cardiovascular  & 30 & 28,154& 2.51& 0.08 & 16 & 14 \\ 
        \hline
    \end{tabular}
\vspace{-10pt}
\end{table}

Table~\ref{tab:lab_den} provides a summary of the multi-label problems used in this research. For multi-label problems, the notations as per Tsoumakas et al., (2009)~\cite{tsoumakas2009mining} are used, where $L = \{\lambda_j:j=1...q\}$ refers to the finite set of labels and $D = \{(x_i,Y_i),i=1...m\}$ refers to set of multi-label training examples. Here $x_i$ is the feature vector, and $Y_i\subseteq L$ is the set of labels of the $i$-th example. Label cardinality ($LCard$) is the average number of labels of the examples in a dataset, and label density ($LDens$) is cardinality divided by $q$. 
%For single-label classification, LCard will always be $1$, and for multi-label problems, LCard will always be $>1$.
Table~\ref{tab:lab_den} provides the number of labels selected for experiments presented in this paper, with the frequency of occurrences $<1\%$, tail-end labels, and the number of labels $\geq1\%$.

\section{Language Models}

This research mainly focuses on transformer models. Transformers are feed-forward models based on the self-attention mechanism with no recurrence. Self-attention takes into account the context of a word while processing it. Similar to the sequence-to-sequence attention mechanism, self-attention is considered a soft measure where multiple words are considered. Transformer models take all the tokens in the sequence at once in parallel, enabling the capture of long-distance dependencies. Vaswani et al. (2017)~\cite{vaswani2017attention} provides an introduction to the transformer architecture. 

BERT (Bidirectional Encoder Representations from Transformers) \cite{DBLP:journals/corr/abs-1810-04805} is one of the early transformer models that applies bidirectional training of encoders \cite{vaswani2017attention} to language modelling. The 12-layer BERT-base model with a hidden size of 768, 12 self-attention heads, 110M parameter neural network architecture, was pre-trained from scratch on BookCorpus and English Wikipedia. PubMedBERT~\cite{gu2020domain} uses the same architecture, and is domain-specifically pre-trained from scratch using abstracts from PubMed and full-text articles from PubMedCentral to better capture the biomedical language \cite{gu2020domain}. 

BioMed-RoBERTa-base~\cite{domains} is based on the RoBERTa-base \cite{DBLP:journals/corr/abs-1907-11692} architecture. RoBERTa-base, originally trained using 160GB of general domain training data, was further continuously pre-trained using 2.68 million scientific papers from the Semantic Scholar corpus. Gururangan et al. (2020)~\cite{domains} show that BioMED-RoBERTa-base, which was specifically pre-trained on medical text data, outperforms the generically trained RoBERTa-base model on biomedical domain-specific tasks. 

TransformerXL~\cite{dai2019transformer} is an architecture that enables the representation of language beyond a fixed length. It can learn dependency that is longer than recurrent neural networks and vanilla transformers. The Longformer~\cite{beltagy1904longformer} model is designed to handle longer sequences without the limitation of the maximum token size of 512. Longformer reduces the model complexity from quadratic to linear by reformulating the self-attention computation. Compared to Transformer-XL~\cite{dai2019transformer}, Longformer is not restricted to the left-to-right approach of processing documents.

In addition to transformer models, CNNText~\cite{kim2014convolutional} with domain-specific fastText pre-trained 100-dimensional embeddings is used. CNNText combines one-dimensional convolutions with a max-over-time pooling layer and a fully connected layer. The final prediction is made by computing a weighted combination of the pooled values and applying a sigmoid function. A simple architecture of CNNText is presented in Figure \ref{fig:dual}.

CAML~\cite{mullenbach2018explainable} is also used to compare with TransformerXL and other languange models. CAML combines convolution networks with an attention mechanism. Simultaneously, a second module is used to learn embeddings of the descriptions of ICD-9 codes to improve the predictions of less frequent labels and target regularisation. For each word in a given document, word embeddings are concatenated into a matrix, and a one-dimensional convolution layer is used to combine these adjacent embeddings.

\begin{algorithm}[t!]
\caption{Multiple BioMed-Transformer}
	\label{alg:multiple}
	\begin{algorithmic}[1]
		\STATE\textbf{Input:} Fixed length multi-sourced or long document text input with set of labels $Y\subseteq L$,  domain specific pre-trained transformer models $x_i$  with parameters $\theta_{1,2,...,n}$, Linear layer (FC) with $L$ number of output units having $\theta_l$ parameters and loss function Binary-cross-entropy (BCE).
		\FOR {each mini-batch }
	    	\STATE pooled$\_features$ = []
		        \FOR {each document $i$}
    		        \STATE $x_i$ = $\text{BioMed-Transformer(document}_i)$ 
    		        \STATE  pooled$\_$features.append( AVG$\_$POOL($x_i$))
                \ENDFOR
    		\STATE combined$\_$features = CONCATENATE(pooled$\_$features)
    		\STATE drop$\_$output = DROPOUT(combined$\_$features)
    		\STATE output = FC$_{\theta_l}$(drop$\_$output)
    		\STATE $\mathcal{L} = \mathcal{L}_{BCE}$(output, targets)
    		\STATE $\theta = [\theta_1,\theta_2,\theta_3,\ldots,\theta_n ,\theta_l]$
            \STATE $\theta =  \theta -   \nabla_\theta \mathcal{L}	$
 		\ENDFOR

	\end{algorithmic}
	\end{algorithm}
	\vspace{-1em}

\section{Concatenated Language Models}

%This section presents the details of three variations of concatenated language models: multi-CNNText, multi-BioMed-Transformers and CNNText with Transformers. 

\subsection{Multi-BioMed-Transformers}

Multi-BioMed-Transformers use an architecture where two or more domain-specific transformer models are concatenated together to enable the usage of multiple text inputs. Algorithm~\ref{alg:multiple} presents an outline of multi-Bio-Med-Transformer models concatenated together. We explore the options of two to four PubMedBERT models that are concatenated together. See Figure~\ref{fig:dual} for an example of TriplePubMedBERT architecture. Concatenated transformer models enable the processing of longer sequences, where the longer input sequence is split into multiple smaller segments with a maximum  length of 512 tokens. The average length of discharge summaries in MIMIC-III is approximately $1,500$ tokens, hence the choice to concatenate two to four PubMedBERT models. Moreover, as indicated in Section~\ref{sec:data}, MIMIC-III contains text from other categories, such as \textsf{ecg} and \textsf{rad}. Multi-BioMed-Transformers provides the option to explore using these other available texts as additional input text. 
%Figure~\ref{fig:dual} presents an example architecture of three PubMedBERT models concatenated together.

\subsection{Multi-CNNText}

Multi-CNNText adopts the same idea as multi-BioMed-Transformers, where two or more CNNText models are concatenated together. Figure~\ref{fig:dual} presents an example of DualCNNText where two CNNText models are concatenated together. Although CNNText can handle longer sequence length as input text, concatenating multiple CNNText models provides the option of using input text from different categories such as ECG and radiology, as mentioned before, as the features of different categories can be captured separately. 

\subsection{CNNText with Transformers}

\begin{figure}[t!]
\centering
\begin{subfigure}[b]{\textwidth}
        \centering
        \includegraphics[width=0.55\textwidth]{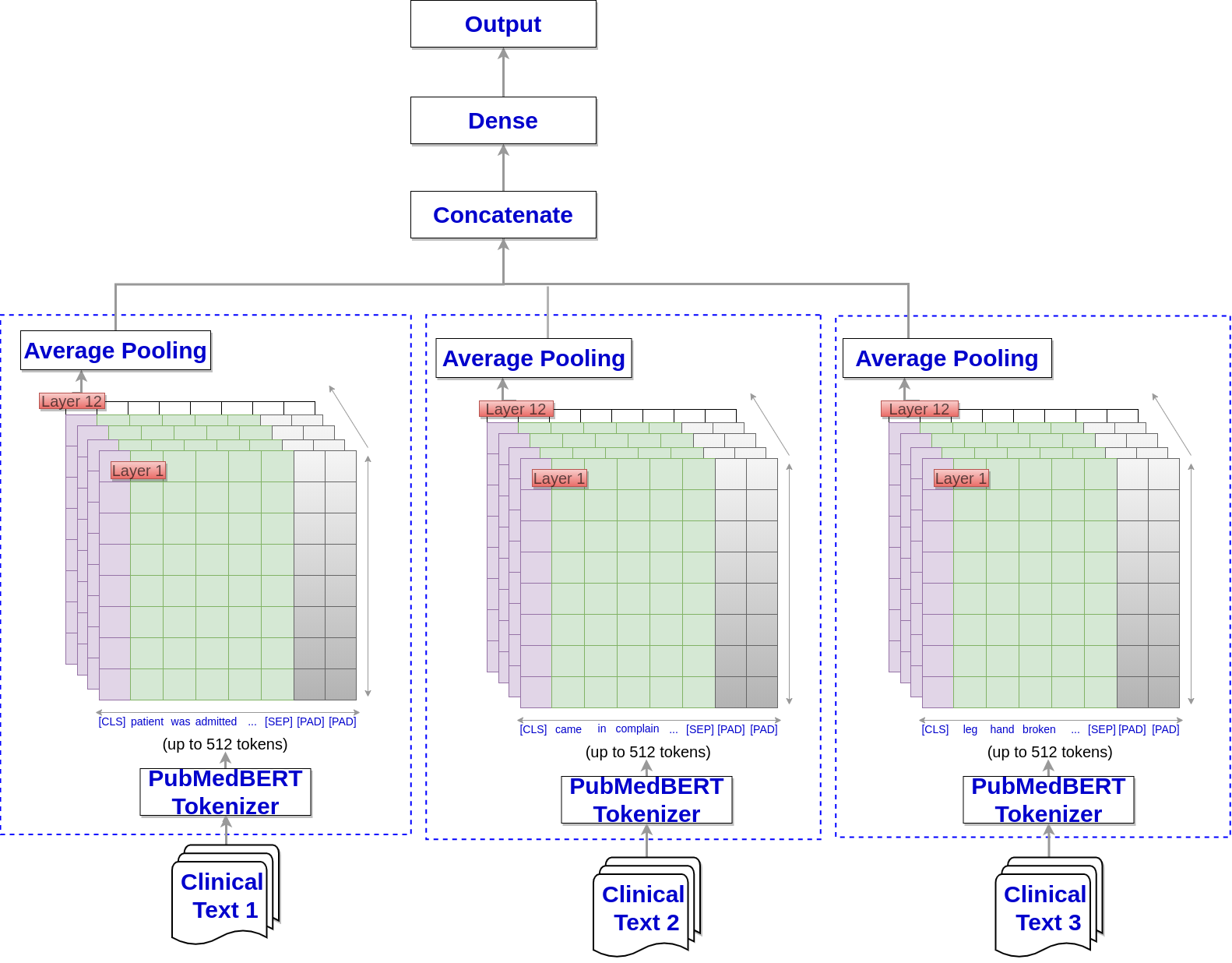}
         \caption{TriplePubMedBERT architecture.}
    \end{subfigure}
    \begin{subfigure}[b]{\textwidth}
        \centering
        \includegraphics[width=0.5\textwidth]{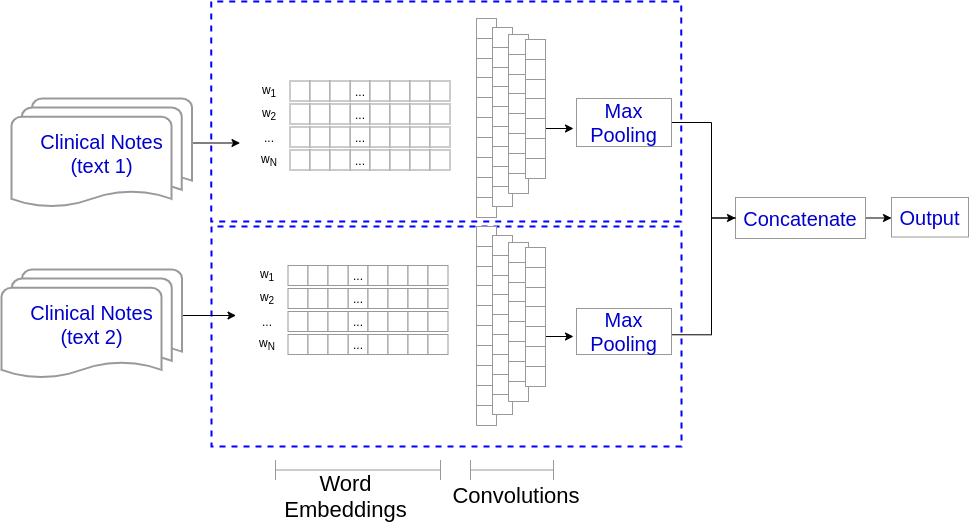}
         \caption{DualCNNText architecture.}
    \end{subfigure}
        \begin{subfigure}[b]{\textwidth}
        \centering
        \includegraphics[width=0.5\textwidth]{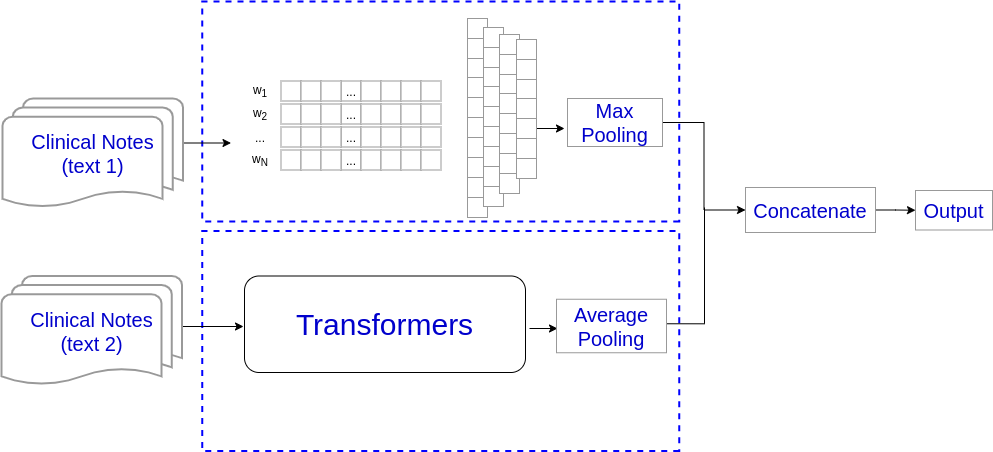}
         \caption{CNNText with Transformer architecture}
    \end{subfigure}
\caption{Concatenated Language Model architectures. }
\vspace{-1.5em}
\label{fig:dual}
\end{figure}

The third variation is combining CNNText with transformers (see Figure~\ref{fig:dual}). Although many variations are possible, this research only considers a couple of variations. BERT-base and PubMedBERT are the two transformers that are used with CNNText. However, variations, such as embeddings dimensions, and multiple transformer models can be used for CNNText. It is also important to point out that CNNText is just one possible choice, and there are many other deep learning models that could be used instead of CNNtext.

\section{Experiments}

We present overall micro and macro F1 scores and individual label F1 scores for the multi-label problems outlined in Table~\ref{tab:lab_den}.  Critical difference plots are presented as supportive statistical analysis. The Nemenyi posthoc test (95\% confidence level) identifies statistical differences between learning methods. CD graphs show the average ranking of individual F1 scores obtained using various language models. The lower the rank, the better it is. The difference in average ranking is statistically significant if there is no bold line connecting the two settings. All experimental results are obtained from a random seeds training-testing scheme and averaged over three runs. The variation of these three independent runs are within a range of $\pm 0.015$.  We explore several different transformer models and compare the performance to concatenated BioMed-Transformers. 

Transformer implementations are based on the open-source PyTorch transformer repository.\footnote{https://github.com/huggingface/transformers} Transformer models are fine-tuned on all layers without freezing. For the optimiser, we use Adam~\cite{kingma2014adam} with learning rates between 9e-6, and 1e-5. Training batch sizes were varied between 1 and 16.  A non-linear sigmoid function $f(z) = \frac{1}{1+e^{-z}}$, with a range of 0 to 1 is used as the activation function. Binary-cross-entropy~\cite{cox1958regression} loss, $Loss_{BCE}(X,y) = -\sum_{l=1}^{L}(y_l log(\hat{y}_l)+(1-y_l)log(1-\hat{y}_l))$, over each label is used for multi-label classification. Domain-specific fastText embeddings~\cite{yogarajan2020,yogarajan2020seeing} of a 100-dimensional skipgram model are used for neural networks.\footnote{Our source code can be obtained from: \\ \url{https://github.com/vithyayogarajan/Medical-Domain-Specific-Language-Models/tree/main/Concatenated-Language-Models-Multi-label}} 

\section{Results}

Results are presented in three parts. First we present the overall performance of the language models, followed by SOTA comparison, and finally we present tail-end performance.

\subsection{Overall performance}\label{sec:overall}

\begin{figure}[b!]
    \includegraphics[width=1\textwidth,height=4cm]{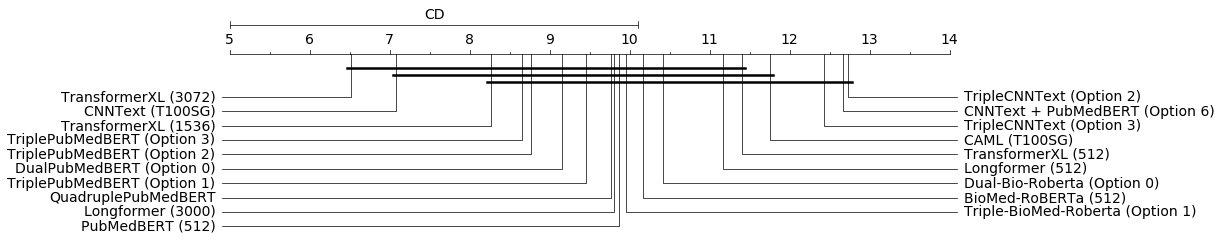}
    \caption[Critical difference plots.]{Critical difference plots. Nemenyi post-hoc test (95\% confidence level), identifying statistical differences between language models for cardiovascular disease presented in Table~\ref{tab:cardio28154_dual}.}
    \label{fig:cardio_dual_cd}
\end{figure}

We present an extensive comparison across models for cardiovascular disease, followed by selected results for other multi-label problems. Table~\ref{tab:cardio28154_dual} presents the results for various language model variations for cardiovascular disease, using MIMIC-III data with 28,154 hospital admissions of patients and 30 labels. Multi-PubMedBERT and multi-BioMed-RoBERTa show a consistent improvement of 3\% to 7\% in micro-F1 scores over single PubMedBERT and BioMed-RoBERTa, respectively. The macro-F1 score of TriplePubMedBERT option is better than other language models presented with at least 3\% improvement, except for TransformerXL with 3,072 tokens. Macro F1 scores of multi-CNNText and CNNText with transformers perform poorly compared to all other language models presented. For cardiovascular disease, incorporating \textsf{ecg} and \textsf{rad} does show some improved overall results, especially with TriplePubMedBERT options. Critical difference plots for individual label F1 scores obtained using various language models in Table~\ref{tab:cardio28154_dual} are presented in Figure~\ref{fig:cardio_dual_cd}. Both Table~\ref{tab:cardio28154_dual} and Figure~\ref{fig:cardio_dual_cd} show that TransformerXL with \textsf{dis} 3,072 tokens is the best option. However, multi-BioMed-Transformers show improvements, especially when compared to single-BioMed-Transformers.

\begin{table}[t!]
        \caption{Comparison of micro-F1, macro-F1 of cardiovascular disease among various language models and input text for MIMIC-III data. Input text options include the maximum sequence length and reference to the options. Bold is used to indicate the best results for each grouping in the table, and underline is used for overall best results. Results are averaged over three runs. }
    \label{tab:cardio28154_dual}
    \centering
     \resizebox{\linewidth}{!}{
    \begin{tabular}{llcc}
    \hline
Neural Network Details & Input Text Options & Micro-F1 & Macro-F1 \\ \hline
BioMed-RoBERTa& dis 512 &0.69 &0.30 \\
PubMedBERT  &  dis  512&	0.70 &	{0.30} \\
TransformerXL &  dis   1,536 & 0.75 & 0.28  \\
TransformerXL &  dis   3,072 & \underline{\textbf{0.78}} & \underline{\textbf{0.32}}  \\
Longformer &  dis   3,000 & 0.74&0.30  \\
CAML (T100SG) & dis  3,000 & 	{{0.77}}	& 0.24  \\
\hdashline\noalign{\smallskip}
Dual-Bio-RoBERTa & Option 0: 512  &0.72 & 0.28  \\ 
DualPubMedBERT & Option 0: 512  &	0.72 & 	{0.30}  \\
Triple-BioMed-RoBERTa &Option 1: 512  & 0.72 & 0.29  \\
TriplePubMedBERT &Option 1:  512   &  0.73 & 0.29  \\
TriplePubMedBERT &Option 2: 512  &	{0.73} &	\textbf{0.31}  \\
TriplePubMedBERT &Option 3: 512  &	{0.73} &	{0.30}  \\
QuadruplePubMedBERT & Option 4: 512 	& \textbf{0.74} &	0.28  \\
\hdashline\noalign{\smallskip}
CNNText (T100SG) &dis 512&0.72 &0.23  \\
CNNText (T100SG) &  dis   3,000 &	0.74 &	\textbf{0.30}  \\
DualCNNText (T100SG) & Option 0:  1,000&0.73&0.22  \\
TripleCNNText (T100SG) &Option 2: 1,000 &0.74&0.24  \\
TripleCNNText (T100SG) &Option 3: 1,000& \textbf{0.75}& {0.25}  \\
QuadrupleCNNText (T100SG) & Option 4: 1,000 & 0.74&0.22  \\
\hdashline\noalign{\smallskip}
CNNText (T100SG) + BERT-base &  Option 6: dis 3,000 + ecg 512  &0.75 &0.20  \\
CNNText (T100SG) + PubMedBERT &Option 6:  dis 3,000 + ecg 512  &\textbf{0.76} &\textbf{0.22}  \\
CNNText (T100SG) + PubMedBERT &Option 7:  dis 3,000 + rad 512  & 0.75 &0.21  \\
\hline
    \end{tabular}}

\end{table}

\begin{table}[t!]
    \centering
    \caption{Comparison of micro-F1, macro-F1 of cardiovascular disease, systemic fungal or bacterial infection, levels 2 and 3 of ICD-9 codes  among various language models. Time required per epoch\protect\footnotemark[3] of systemic fungal or bacterial infection and MIMIC-III Level 3 is also presented. Input text options are included for reference. Bold is used to indicate the best results among the groups except for time where it is the lowest time, and underline is used for overall best results. Published results are also presented for direct comparison. Results are averaged over three runs. }
    \label{tab:dual}
      \resizebox{\linewidth}{!}{
    \begin{tabular}{llrrr:rrr}
    \hline \noalign{\smallskip}
& & \multicolumn{2}{c}{COVID-19} & &  \multicolumn{3}{:c}{Fungal or Bacterial} \\
Transformers & Input Text  &Micro-F1 & Macro-F1 &  \quad \quad & Micro-F1 & Macro-F1 & Time (epoch) \\ 
 \noalign{\smallskip}\hline\noalign{\smallskip}
BioMed-RoBERTa&  dis 512  & 0.53~\cite{yogarajan2021predicting}& 0.45~\cite{yogarajan2021predicting}&& 0.45&0.39 & {2,554 sec} \\
PubMedBERT  &  dis 512 & 0.54~\cite{yogarajan2021predicting}	 &0.48~\cite{yogarajan2021predicting}	 & &0.48 &0.39 & 2,940 sec \\
TransformerXL &  dis 512   & 0.51 &0.45  & &0.47 &0.39 &2,921 sec \\
TransformerXL &  dis 3,072 & \underline{\textbf{0.65}}~\cite{yogarajan2021predicting}  & \underline{\textbf{0.51}}~\cite{yogarajan2021predicting}  &   &\underline{\textbf{0.64}} &\underline{\textbf{0.46}}  &{43,200 sec} \\
Longformer &  dis  3,000 &0.58~\cite{yogarajan2021predicting} &0.50~\cite{yogarajan2021predicting} & &0.58 &0.43 & 13,500 sec \\
CAML (T100SG)& dis 3,000 &0.61~\cite{yogarajan2021predicting} &0.40~\cite{yogarajan2021predicting} & & {0.62}&0.38 & \textbf{47 sec} \\
& & & & & & & \\
DualPubMedBERT & Option 0: 512 &	\textbf{0.58} &\textbf{0.49} 	 & & \textbf{0.57} & \textbf{0.43} & 4,020 sec\\
TriplePubMedBERT & Option 1: 512  & 0.54  &0.46  &  &0.56 &0.40 & 5,580 sec  \\
TriplePubMedBERT & Option 2: 512  &	 -&	- & & 0.54& 0.39 & 5,580 sec\\
TriplePubMedBERT & Option 3: 512    &	- &- & &0.54 &0.39  & 5,580 sec\\
QuadruplePubMedBERT & Option 4: 512 	& - &	- & &0.54 &0.40 & 7,080 sec   \\
QuadruplePubMedBERT & Option 5: 512   	&  0.52&	0.46 & &\textbf{0.57} &0.40 & 7,080 sec \\

\noalign{\smallskip}\hline\noalign{\smallskip}
 %   \end{tabular}}
% \vspace{-8pt}  
%\end{table}

%\end{center}
%\vspace{+4pt}
%\end{table}
%\begin{table}[t!]
 %   \centering
 %     \caption{Comparison of micro-F1, macro-F1 of levels 2 and 3 of ICD-9 codes among various language models. Bold is used to indicate the best results for each grouping in the table, and underline is used for overall best results. Results are averaged over three runs.}\label{tab:level2_3_dual}
%     \resizebox{\linewidth}{!}{
  %  \begin{tabular}{llrrrrrr}
    %   \hline\noalign{\smallskip}
 &   & \multicolumn{2}{c}{MIMIC-III Level 2 codes} &  &  \multicolumn{3}{c}{MIMIC-III Level 3 codes} \\  
Transformers & Input Text  &Micro-F1 & Macro-F1 &  \quad \quad & Micro-F1 & Macro-F1 & Time (epoch) \\ 
\noalign{\smallskip}\hline\noalign{\smallskip}
PubMedBERT &  dis 512 & 0.65~\cite{yogarajan2021multilabel} & 0.41~\cite{yogarajan2021multilabel} &  & 0.55 & 0.18 & \textbf{3,393 sec}\\
BioMed-RoBERTa & dis 512 & 0.64~\cite{yogarajan2021multilabel} & 0.40~\cite{yogarajan2021multilabel} & & 0.53 & 0.18 &  4,877 sec \\
TransformerXL &  dis 3,072  & \underline{\textbf{0.73}} & \underline{\textbf{0.46}} & & -&- &- \\
Longformer & dis 3,000  & {{0.72}} & {{0.45}} & &  {{0.62}} & 0.19 & {16,889 sec} \\
CAML (T100SG)& dis 3,000 & {{0.72}} & 0.43 & &  \underline{\textbf{0.64}} &  \underline{\textbf{0.26}} & \textbf{64 sec}\\
& & & & & & & \\
DualPubMedBERT & Option 0: 512  &\textbf{0.68} &{\textbf{0.45}} & &\textbf{0.57} &\underline{\textbf{0.20}} & 4,750 sec\\
DualBioMed-RoBERTa &  Option 0: 512   &0.66 &0.43 &  &0.56 &0.19 & 6,842 sec\\
TriplePubMedBERT & Option 1: 512   & 0.66 &0.43 & & - & -  & - \\ 
 \noalign{\smallskip}\hline\noalign{\smallskip}
    \end{tabular}}
    \vspace{-8pt}
    \end{table}

\begin{figure}[t!]
    \begin{subfigure}[b]{\textwidth}
        \centering
        \includegraphics[width=\textwidth,height=2.8cm]{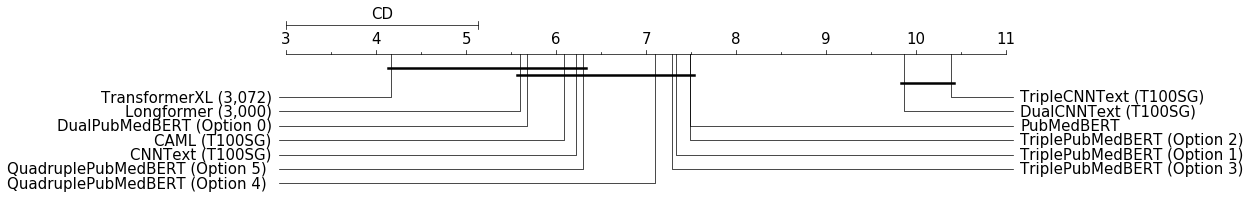}
        \caption{Task: Systemic fungal or bacterial infections}
    \end{subfigure}
    \begin{subfigure}[b]{\textwidth}
        \centering
        \includegraphics[width=\textwidth,height=2.8cm]{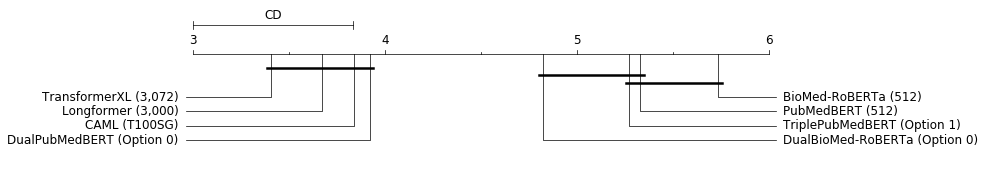}
        \caption{Task: MIMIC-III Level 2 codes}
    \end{subfigure}
    \caption{Critical difference plots. Nemenyi post-hoc test (95\% confidence level), identifying statistical differences between language models in Table~\ref{tab:dual}, where critical difference is calculated for individual label F1 scores. }
    \label{fig:cd}
    \vspace{-4pt}
\end{figure}

Table~\ref{tab:dual} presents micro and macro F1 scores for various language model variations for COVID-19 patient shielding and systemic fungal or bacterial infection using MIMIC-III data. For systemic fungal or bacterial infections multi-PubMedBERT show improvements of 12\% to 19 \% in micro-F1, and 2\% to 10\% in macro-F1 scores over single PubMedBERT, except for TriplePubMedBERT with \textsf{rad} and \textsf{ecg}, where the macro-F1 score is on par with single PubMedBERT. Contrary to the case of cardiovascular disease, here the additional inputs of \textsf{ecg} and \textsf{rad} do not result in better performance. 
It is likely that \textsf{ecg} and \textsf{rad} are not that relevant for coding fungal or bacterial infections. Table~\ref{tab:dual} for COVID-19 patient shielding shows TransformerXL with \textsf{dis} 3,072 tokens to be the best option, as observed with the other case studies. DualPubMedBERT show improvements over single PubMedBERT and other variations of multi-PubMedBERT.

All three case studies show that TransformerXL with \textsf{dis} 3,072 tokens is the top performer in terms of predictive performance. However, concatenated BioMed-Transformers show improvements, especially when compared to single BioMed-Transformers. Table~\ref{tab:dual} also presents the time per epoch in seconds for systemic fungal or bacterial infection to provide a direct comparison among the language models. TransformerXL (3,072) requirements are much greater than that of other language models, including multi-PubMedBERT, for example, needing 240 hours (for \textsf{dis} 3,072) when DualPubMedBERT only requires 22 hours.
%For example, if dualPubMedBERT is directly compared to TransformerXL (3,072), the time required per epoch is 10x more, i.e. for the experimental results presented in this paper, TransformerXL (3,072) requires 240 hours or ten days for training, compared to dualPubMedBERT which only requires 22 hours.  Table 3 also shows 

Table~\ref{tab:dual} also presents micro and macro F1 scores for levels 2 and 3 of ICD-9 codes using MIMIC-III data. As mentioned above, due to the processing time required by TransformerXL (3,072), we only use Longformer for encoding long documents for ICD-9 level 3. For MIMIC-III Level 2 codes, TransformerXL with \textsf{dis} 3,072 tokens is the top performer. DualPubMedBERT shows improvements in both micro and macro F1 scores by 3\% to 5\% over other PubMedBERT variations, and macro-F1 of DualPubMedBERT and Longformer are equal and only marginally behind TransformerXL. For MIMIC-III Level 3 codes, the macro-F1 score of DualPubMedBERT is better than other transformer models, including Longformer. However, CAML (T100SG) outperforms all variations of transformer models.   

Figure~\ref{fig:cd} presents the critical difference plots for
%selected multi-label problems as supportive statistical analysis for
results presented in Table~\ref{tab:dual}. The Nemenyi posthoc test (95\% confidence level) shows statistical differences between learning methods. 
%CD graph shows the average ranking of individual F1 scores obtained using various language models. The lower the rank, the better it is. The difference in average ranking is statistically significant if there is no bold line connecting the two settings. 
TransformerXL (3,072) and Longformer (3,000) are the overall top performers. However, the difference between them and DualPubMedBERT is not statistically significant.
    
\footnotetext[3]{Average times (in seconds) based on experiments run on 12 core Intel(R) Xeon(R) W-2133 CPU @ 3.60GHz, GPU device GV100GL [Quadro GV100].}   

This section compares the overall performance of multiple language models for MIMIC-III data for the number of labels being 30, 42, 73, 158 and 923. TransformerXL (3,072) consistently outperformed other language models. Multi-CNNText and CNNText with Transformers performed poorly when compared to other language model variations. Hence, only results for cardiovascular disease are presented in this research for the CNNText variations. Multi-BioMed-transformers outperforms single BioMed-Transformers with a more noticeable improvement in micro-F1 scores for cardiovascular disease and systemic fungal or bacterial infections. Due to computational restrictions, only Longformer was used to handle long text sequences for level 3 of ICD-9 codes. DualPubMedBERT macro-F1 score was the same as Longformer and TransformerXL for level 2 ICD-9 codes, and better than Longformer for for level 3 ICD-9 codes with 923 labels.

\subsection{SOTA Results}

Both Tables~\ref{tab:cardio28154_dual} and \ref{tab:dual} and Figures~\ref{fig:cardio_dual_cd} and \ref{fig:cd} show that TransformerXL outperforms CAML across all multi-label problems for predicting medical codes. In addition, there are other language models, including concatenated models, that perform on par with or above CAML, especially when macro-F1 scores are compared. 
\begin{table}[t!]
      \caption{Overall results for MIMIC-III Top 50 ICD-9 codes. Bold is used to indicate the best results. Published results are presented for direct comparison. Description Regularized-CAML is referred by DR-CAML. Both variations of EffectiveCAN is presented.\protect\footnotemark[4]  Our results are averaged over three runs.}
          \label{tab:top50}
        \centering
    \begin{tabular}{lrr}
    \hline
        Models & Micro-F1 & Macro-F1 \\\hline\noalign{\smallskip}
         CAML~\cite{mullenbach2018explainable} & 0.614&0.532 \\ 
         DR-CAML~\cite{mullenbach2018explainable} & 0.633&0.576 \\
          EffectiveCAN (Sum-pooling attention)~\cite{liueffective} & 0.702 & 0.644 \\
         EffectiveCAN (Multi-layer attention)~\cite{liueffective} & 0.717 & 0.668 \\
         \noalign{\smallskip}\hdashline\noalign{\smallskip}
         DualPubMedBERT (Option 0: 512) &0.640 &0.576 \\
        TriplePubMedBERT (Option 1: 512) &0.641 &0.583 \\
        Longformer (3,000) & 0.703&0.654 \\
        TransformerXL (3,072) & \textbf{0.723}&\textbf{0.677} \\
        \noalign{\smallskip}\hline
    \end{tabular}
\end{table}

\footnotetext[4]{See Liu et al (2021)~\cite{liueffective} for details of EffectiveCAN variations and architectures.}

Table~\ref{tab:top50} provides the overall micro and macro F1 scores of the most frequent 50 ICD-9 codes in MIMIC III with discharge summary. In this particular case, for direct comparison, the labels and input data are all matched to the exact specifications of the compared published methods. This is the only section in this research where Top 50 ICD-9 codes are used for experimental evaluations. Evidently TransformerXL (3,072) with a learning rate of 1e-5 presents new SOTA results.

\begin{figure}[tph!]
    \centering
    \includegraphics[width=0.95\textwidth]{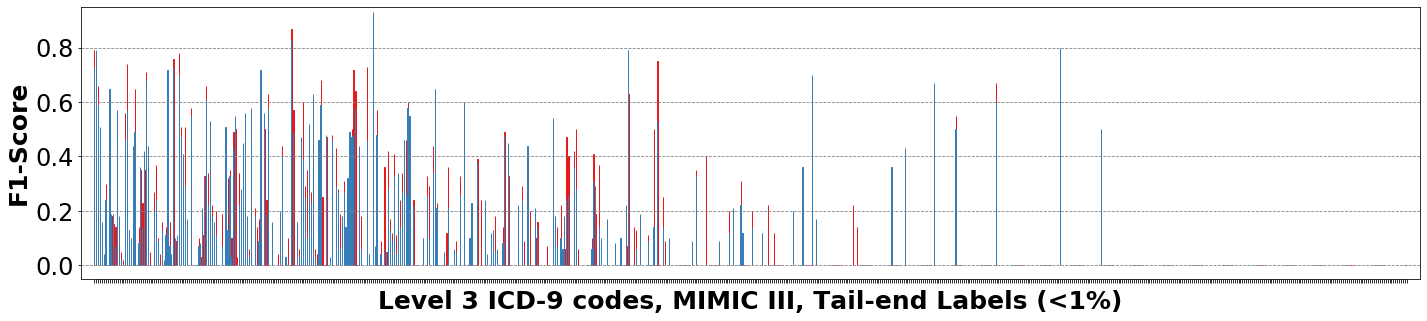}\\
    \includegraphics[width=0.95\textwidth]{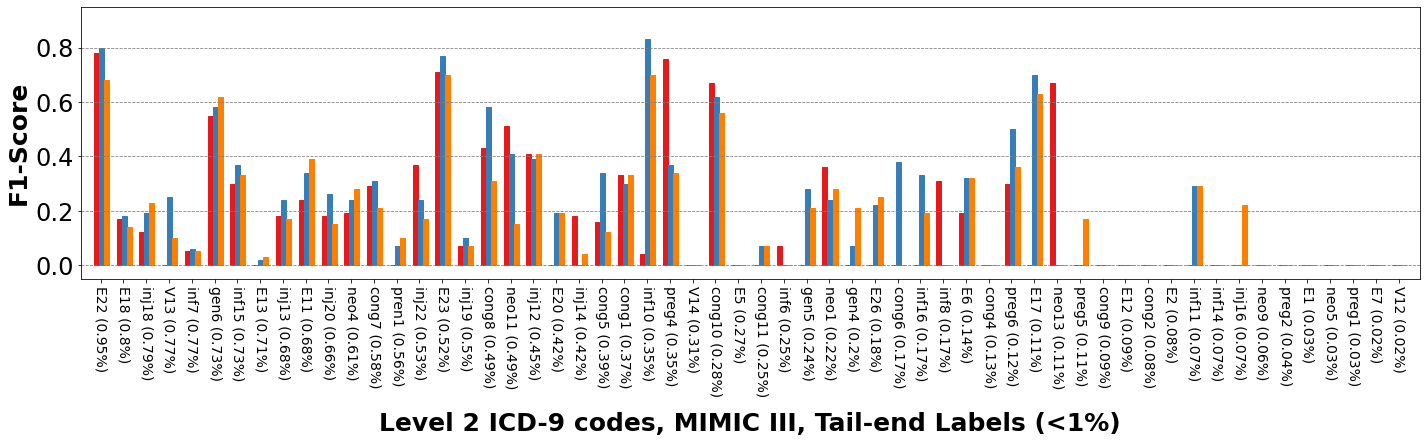} \\
      \includegraphics[width=0.95\textwidth]{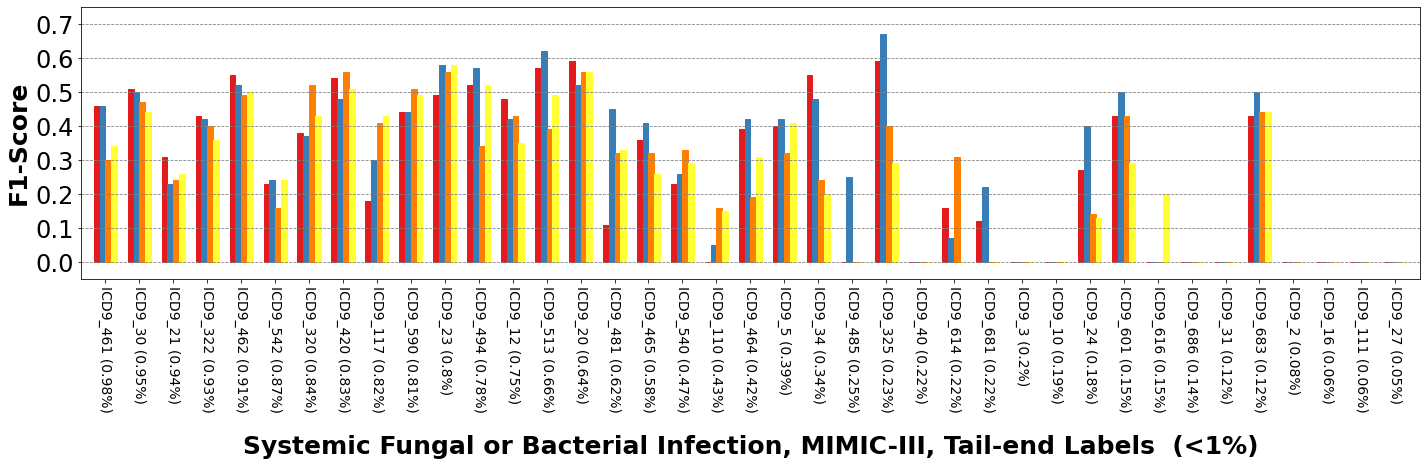}\\
       \includegraphics[width=0.95\textwidth]{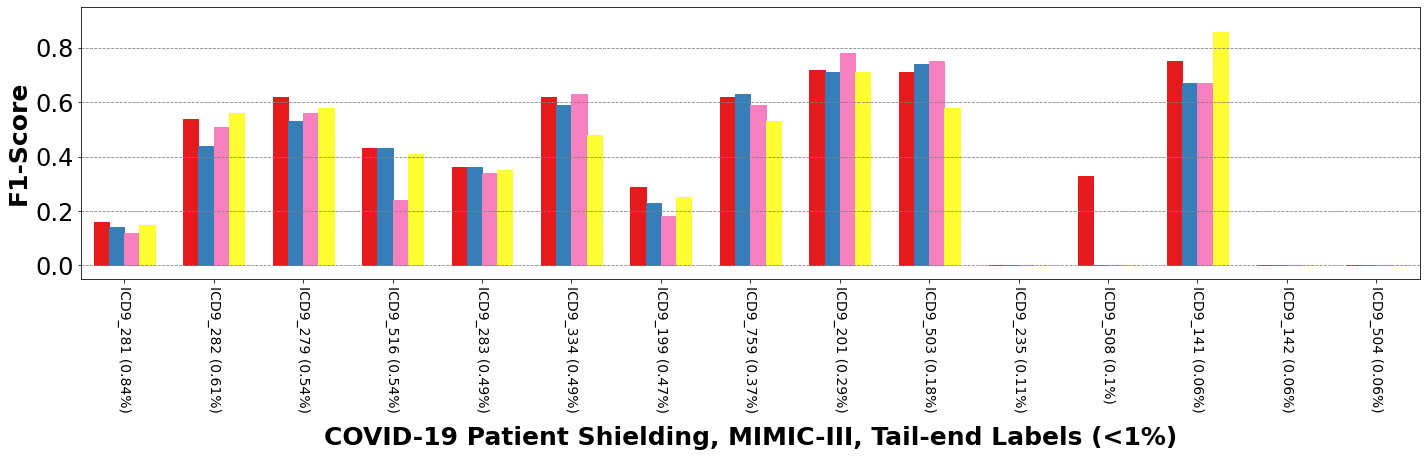}\\
        \includegraphics[width=0.95\textwidth]{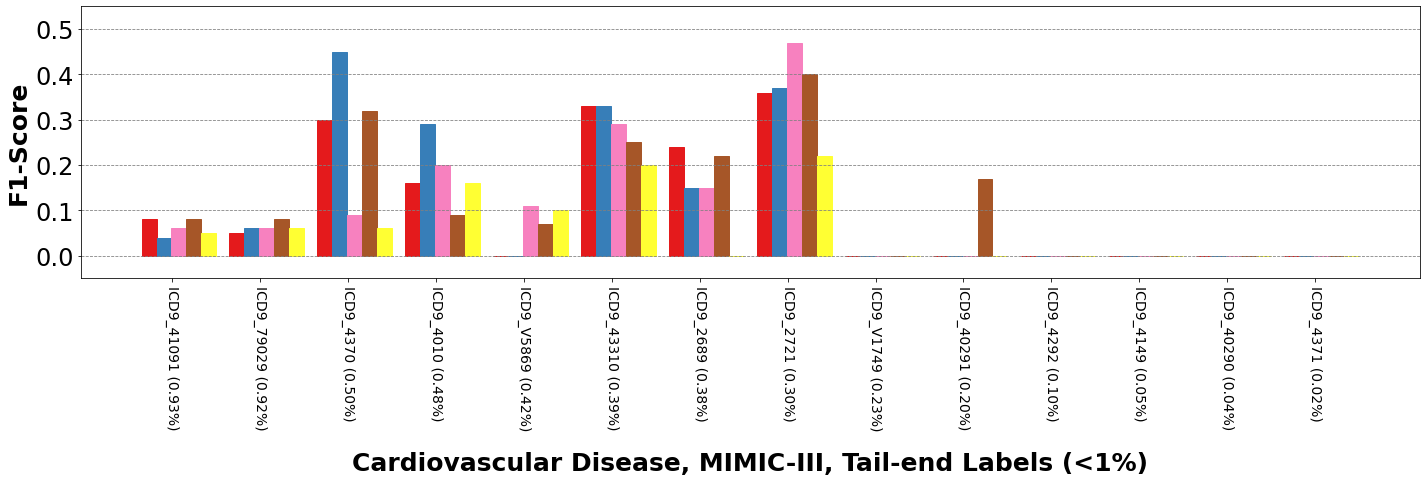}\\
         \includegraphics[width=0.85\textwidth]{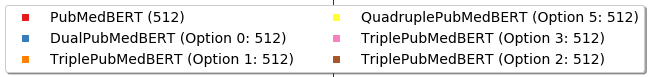}\\
    \caption{F1-score for tail-end labels (frequency $< 1\%$), where single PubMedBERT (in red) is compared with multi-BioMed-transformer models. }
    \label{fig:tail_f1}
    \vspace{-4pt}
\end{figure}

\subsection{Tail-end Labels}

This section presents a comparison of individual label F1 scores for the multi-label problems presented in Section~\ref{sec:overall}. The focus here is on showing the differences and the improvements in F1 scores of tail-end labels with multi-BioMed-Transformers compared to single transformer models including Longformer and TransformerXL. Table~\ref{tab:lab_den} presents the number of labels with frequency $\geq$ 1\% , and tail-end labels (with label frequency $<$ 1\%). 

Figure~\ref{fig:tail_f1} presents tail end F1 scores across all five multi-label problems. With the exception of a few specific labels, including \textsf{ICD-9 code 508} for COVID-19 patient shielding problem, in general F1-scores of concatenated Bio-Med-Transformers for tail-end labels are consistently better. This improvement is more evident for long tail-end cases such as levels 2 and 3 of ICD-9 codes where there is also an improvement in the number of labels with F1-score $\neq$ 0.

\begin{figure}[t!]
    \centering
    \includegraphics[width=0.95\textwidth]{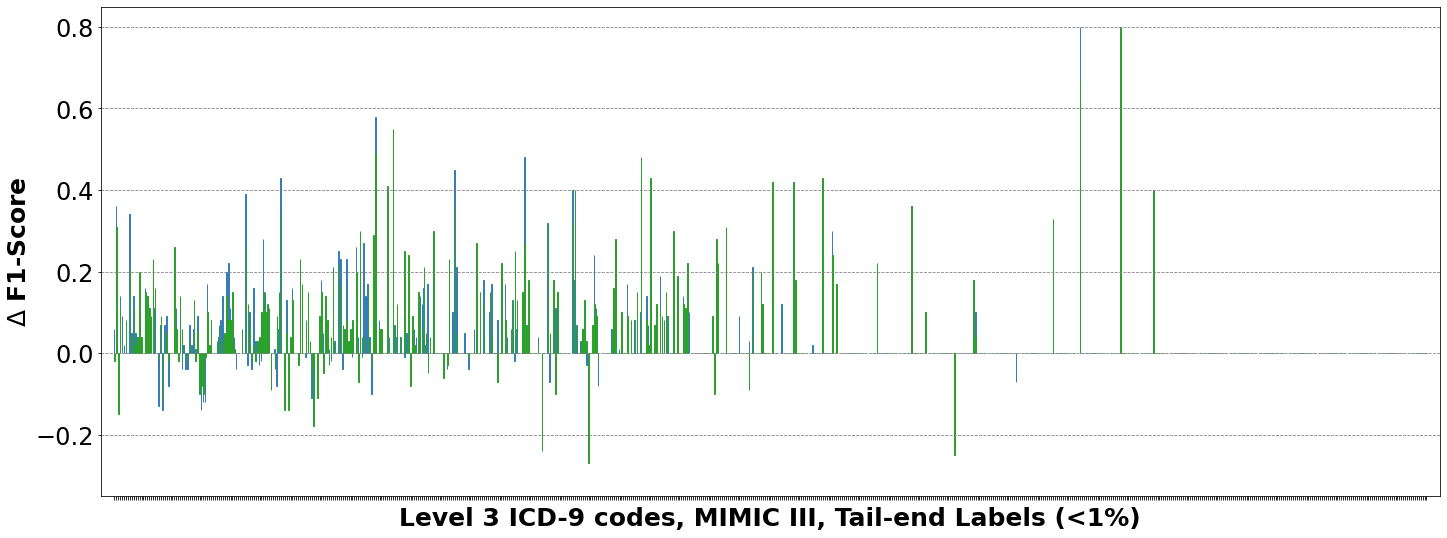}\\
    \includegraphics[width=0.95\textwidth]{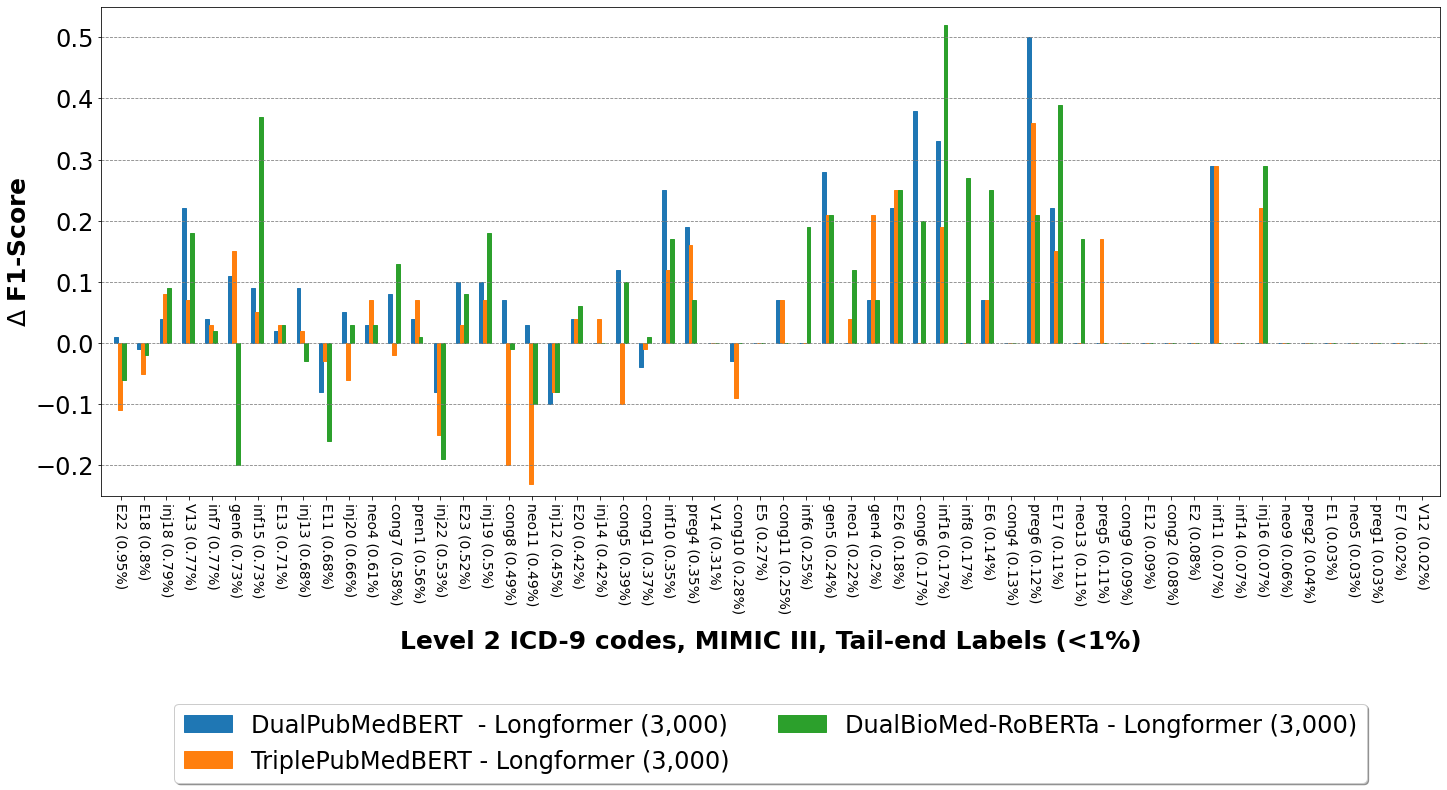}
    \caption{MIMIC-III Level 2 and 3 codes, where the difference between F1 scores of dual/triple language model variations and Longformer (3,000 tokens) is presented. %Top plot: Label frequency $\geq 1\%$ where labels are ordered based on frequency with most frequent label at the left of the plot. 
    Tail-end labels (frequency $< 1\%$) - labels ordered based on frequency with the least frequent label at the right end. Legend is presented for reference. Negative values indicate better F1 scores for Longformer (3,000).  }
    \label{fig:level2_tail_long}
    \vspace{-4pt}
\end{figure}

Tables~\ref{tab:cardio28154_dual} and \ref{tab:dual} show the overall performance of Longformer and TransformerXL in general are better, especially when compared to single Bio-Med-Transformers. To analyse the difference in tail-end label F1 scores we also present the actual difference between F1-scores by calculating the differences of F1 scores between a particular language model variation and Longformer or TransformerXL, hence, negative values indicate Longformer or TransformerXL have a better F1 score.

Figure~\ref{fig:level2_tail_long} presents the difference in F1 scores for level 2 and 3 ICD-9 codes for the following combinations: dualPubMedBERT - Longformer (3,000), triplePubMedBERT - Longformer (3,000) and dualBioMed-RoBERTa - Longformer (3,000). Due to space restrictions only tail-end labels are presented. However, it is important to note that for frequent labels, the F1 scores of Longformer are on par or better than the other three models. Smaller differences in F1 scores are noticed among the most frequent labels, and occasional dual and triple models perform slightly better than Longformer for specific labels. In general, Longformer has the most wins over other models for label frequency $\geq 1\%$. This pattern is reversed for tail-end labels, with Longformer losing more to the dual and triple models where the difference in F1 scores is noted. For some tail-end labels, these differences are noticeably higher than other labels. Level 3 contains 923 labels, with more than 650 labels being infrequent. Figure~\ref{fig:level2_tail_long} also shows Longformer losing more to dual transformers at tail-end labels.

\begin{table}[t!]
    \centering
     \caption{The number of wins, draws and losses of concatenated language models compared to Longformer (LF) and TransformerXL (TXL) for systemic fungal or bacterial infections and levels 2 and 3 of ICD-9 codes.}
    \label{tab:fun_win}
    \begin{tabular}{lrrrrrr}
     \hline
 \multicolumn{7}{c}{\textbf{Systemic fungal or bacterial infections, 73 labels}} \\
  \noalign{\smallskip} \hline
    Models  & \multicolumn{3}{c}{Freq $\geq$ 1\%} &  \multicolumn{3}{c}{ Freq $<$ 1\%} \\
            &  wins & draws & losses & \qquad wins & draws & losses \\
  \noalign{\smallskip} \hline \noalign{\smallskip}
PubMedBERT -TXL&1&1&32&11&12&16\\
DualPubMedBERT - TXL&3&3&28&15&9&15\\
TriplePubMedBERT - TXL &4&0&30&8&13&18\\
QuadruplePubMedBERT - TXL &3&0&31&10&12&17\\
\noalign{\smallskip}  \noalign{\smallskip}  
PubMedBERT - LF&4&3&27&13&10&16\\
DualPubMedBERT - LF&12&0&21&14&13&12\\
TriplePubMedBERT - LF&9&2&23&12&12&15\\
QuadruplePubMedBERT - LF&15&1&18&10&10&19\\
\noalign{\smallskip} \hline \noalign{\smallskip}
 \multicolumn{7}{c}{\textbf{MIMIC-III Level 2 codes, 158 labels}} \\
  \noalign{\smallskip} \hline \noalign{\smallskip}
         DualPubMedBERT - TXL & 26 &0 &74 & 26 &16& 16  \\
         DualBioMed-RoBERTa - TXL  & 25 &0 &75 & 28 &16&14   \\
         TriplePubMedBERT - TXL  & 20 & 1& 79& 27 &14 & 17  \\
         \noalign{\smallskip}  \noalign{\smallskip}  
         DualPubMedBERT - LF & {19} & 11 & 70 & {30} &22 & 6  \\
         DualBioMed-RoBERTa - LF &{{12}} &5 &83  & {{28}} &21 & 9  \\
         TriplePubMedBERT - LF & {{13}} &2 & 85 & {{26}} &20 & 12  \\
\noalign{\smallskip}   \hline \noalign{\smallskip} 
 \multicolumn{7}{c}{\textbf{MIMIC-III Level 3 codes, 923 labels}} \\
\hline \noalign{\smallskip}
         DualPubMedBERT -LF &  {{92}} &17 & 135 & {{160}} & 476 & 43  \\
         DualBioMed-RoBERTa -LF & {{83}} & 21& 144  &{{181}} &454 & 40  \\ 
   \noalign{\smallskip}    \hline
    \end{tabular}
     \vspace{-6pt}  
\end{table}

Table~\ref{tab:fun_win} presents the number of per-label wins, draws, and losses for levels 2 and 3 ICD-9 codes, and fungal or bacterial infections. For multi-label problems, F1-scores of many infrequent labels are zero. This observation is also evident in Figures~\ref{fig:tail_f1} and \ref{fig:level2_tail_long}. To quantify the observations, differences of F1 scores are presented as wins, draws and losses. For most cases, draws are where the F1 scores are zero. We acknowledge that there is a need for further analysis to understand the behaviour observed in Table~\ref{tab:fun_win}.  

As observed in Figure~\ref{fig:level2_tail_long} for tail-end labels, more wins are observed for concatenated models. DualPubMedBERT is the best performing option with the fewest losses among the more frequent label groups and most wins among the tail-end labels. For MIMIC-III Level 3 codes, the results in Table~\ref{tab:fun_win} show Longformer losing more to dual transformers at tail-end labels. For frequent labels for systemic fungal or bacterial infection, the F1 scores of TransformerXL are consistently better than the PubMedBERT variations and a clear winner. For infrequent labels, multi-PubMedBERT variations perform better than TransformerXL for many labels.

\section{Discussion}

We presented concatenated domain-specific language model variations to improve the overall performance of the many infrequent labels in multi-label problems with long input sequences. Although TransformerXL and Longformer can encode long sequences, and in general, TransformerXL outperforms other models setting new SOTA results, the required computational resources are prohibitive. Concatenated PubMedBERT models outperformed single BioMed-Transformers. There was a noticeable improvement in micro-F1 for multi-BioMed-transformers with cardiovascular disease and systemic fungal or bacterial infection. For larger multi-label problems, dualPubMedBERT, TransformerXL and Longformer achieve the same macro-F1 for MIMIC-III Level 2, but dualPubMedBERT wins for MIMIC-III Level 3.

We also study the impact on predictive performance for less frequent labels. Label frequency is highly biased by the hospital/department the data were collected. If the data were from a fertility ward, the label frequency of pregnancy-related medical codes would be high, while for the cardiovascular ward this may not be the case. However, only being able to predict highly frequent labels well poses risks to a patient's health and well being. Hence, this research also compared individual label F1 scores for multi-label problems focusing on tail-end labels. For larger multi-label problems with long tail-end labels, such as level 2 and 3 ICD-9 codes, multi-BioMed-transformers had more wins than Longformer and TransformerXL. 

This provided experimental evidence shows that, with fewer resources, concatenated BioMed-Transformers can improve overall micro and macro F1 scores for multi-label problems with long medical text. In addition, for multi-label problems with many tail-end labels, multi-BioMed-Transformers outperform other language models when F1 scores of tail-end labels are compared directly. 

There are many avenues of research that arise directly from this research. If processing time or resources are not an issue, then continuous training of TransformerXL and Longformer on health-related data might improve prediction accuracy possibly even for tail-end labels. Concatenating TransformerXL or Longformer is also a possibility. ICD-9 codes have a tree-like hierarchy nature. Hence, predicting ICD-9 codes as a hierarchical multi-label classification problem using transformers to encode medical text is another relevant avenue to explore.

\bibliographystyle{spmpsci}  
\bibliography{references}

\end{document}